\pdfoutput=1

\documentclass[11pt]{article}

\usepackage[final]{acl}

\usepackage{times}
\usepackage{latexsym}

\usepackage[T1]{fontenc}

\usepackage[utf8]{inputenc}

\usepackage{microtype}

\usepackage{inconsolata}

\usepackage{graphicx}

\title{Can Language Models Take A Hint? Prompting for Controllable Contextualized Commonsense Inference}

\author{
Pedro Colon-Hernandez \\ Apple\footnotemark
        \And
        Nanxi Liu$^\ddagger$ \\ Tufts\footnotemark  
        \And
        Chelsea Joe$^\ddagger$ \\ Dartmouth College 
        \And
        Peter Chin \\ Dartmouth College  
        \AND
        Claire Yin \\ MIT CSAIL 
        \And
        Henry Lieberman \\ MIT CSAIL 
        \And
        Yida Xin \\ Boston University  
        \AND
        Cynthia Breazeal \\ MIT Media Lab 
}

\begin{document}
\maketitle
\def\thefootnote{$^\ddagger$}\footnotetext{These authors contributed equally to this work}\def\thefootnote{\arabic{footnote}}

\addtocounter{footnote}{1}
\footnotetext{Work done while at the MIT Media Lab}
\addtocounter{footnote}{1}

\footnotetext{Work done while an undergraduate researcher at Wellesly College}

\begin{abstract}
Generating commonsense assertions within a given story context remains a difficult task for modern language models. Previous research has addressed this problem by aligning commonsense inferences with stories and training language generation models accordingly. One of the challenges is determining which topic or entity in the story should be the focus of an inferred assertion. Prior approaches lack the ability to control specific aspects of the generated assertions. In this work, we introduce "hinting," a data augmentation technique that enhances contextualized commonsense inference. "Hinting" employs a prefix prompting strategy using both hard and soft prompts to guide the inference process. To demonstrate its effectiveness, we apply "hinting" to two contextual commonsense inference datasets: ParaCOMET \cite{Gabriel_PC} and GLUCOSE \cite{mostafazadeh-etal-2020-glucose}, evaluating its impact on both general and context-specific inference. Furthermore, we evaluate "hinting" by incorporating synonyms and antonyms into the hints. Our results show that "hinting" does not compromise the performance of contextual commonsense inference while offering improved controllability.

\end{abstract}

\section{Introduction}
The task of Contextual or Discourse-Aware Commonsense Inference, which consists of generating relevant and coherent commonsense assertions (i.e. facts) for a certain sentence in a story context, while easy for humans, remains challenging for machines \cite{Gabriel_PC}. Within this task, we define an \textit{assertion} as a tuple that contains {a subject}\footnote{We note that here we utilize the term "subject" as a part of the relation tuple, and it is not necessarily "subject" in a grammatical sense. In the case of ATOMIC, a subject could be a sentence describing an event that causes another event or a reaction, whereas in ConceptNet it could be a single concept.}, {a relation}, and {an object} (e.g.,\textit{ {a dog}, {is a}, {animal}}), similar to a subject-verb-object triple.  An assertion in this task is a contextually \textit{specific} fact or \textit{generally} applicable templated fact, that can be inferred from a sentence within a given story context. 

Automated systems (such as pre-trained transformer-based language models \cite{devlin-etal-2019-bert,radford2019language}) struggle with generating contextual assertions since there is an implicit assumption that clues for making predictions can always be found explicitly in the text. \cite{da2019cracking,davison2019commonsense,liu2004conceptnet,zhangKoala2021}. This becomes problematic because a model for this task is essentially forced to use knowledge that it may not have seen during pre-training. Additionally, models are forced to guess what to predict about (e.g., what {the subject} of an assertion is), which may lead to decreased performance (e.g., the model generates an assertion about cats when it should have talked about dogs).  Below we give an example of contextual commonsense inference with a story, \textit{a target sentence}, and some corresponding story \textit{specific} and \textit{general} inferences. The story is picked directly from the ROCStories corpus. \cite{mostafazadeh2016corpus}  

\begin{quote}
\tiny

\textbf{Story:} The hockey game was tied up. The {red team} had the puck. They sprinted down the ice. They cracked a shot on goal! \textit{They scored a final goal!}

\textbf{Story Specific Commonsense Inference:} The red team, {is capable of}, {winning the game}

\textbf{General Commonsense Inference:} { Some people \textit{scored a final goal}}, {causes}, {some people to be happy}
\end{quote}

In this example, we can see the aforementioned problems that models have to deal with.  Although it is commonsense that a final goal will lead to a victory for the {red team}, it is not explicitly stated anywhere in the text.  The pre-training of models may include text related to a sudden death goal from sources such as Wikipedia, but the model has to extrapolate that the final goal in this example is a type of sudden death goal and will concede victory to the {red team}. Similarly, although we may want assertions related to the {red team}, the model has to somehow know that it \textit{needs} to infer assertions about this and \textbf{ not} something else.   

Previous attempts have tackled the problem of contextual commonsense inference by constructing datasets of stories aligned with assertions (i.e., an assertion is given for a sentence in a story), either through automated or human-annotated ways and building a model to, given the story and a target sentence, predict part or the whole assertion. A previous attempt to tackle this problem, ParaCOMET \cite{Gabriel_PC}, trained a GPT-2 language model \cite{radford2019language} to infer {an object} of a commonsense assertion tuple from the ATOMIC \cite{sap2019atomic} knowledge base\footnote{ATOMIC is composed of causal assertions, where a certain subject event, causes a certain object event through a given relation.}. They formulate the task as follows. Given a story, a special sentence identifier token, and a specified {relation} type (i.e. dimension of commonsense), the ParaCOMET model must predict the object of a commonsense assertion. 

Another work that tries to approach this is GLUCOSE \cite{mostafazadeh-etal-2020-glucose}.  Here, a dataset is constructed that consists of stories and human annotations for sentences in the stories.  Human annotations provide \textit{specific} and \textit{general} commonsense assertions. The authors use this data set to train a T5 \cite{2020t5} model to generate both types of assertions. The model takes an input sequence in the form of a story, a relation to predict, and a target sentence, and has to predict both the \textit{general} and \textit{specific} assertions that may be present in the target sentence with the given story context.  In both works, the models are expected to infer from the story, a target sentence, and the relation alone. An example of these formulations can be seen in the Appendix.

Recently, there has been work on exploring \textit{prompting} \cite{liu2021pre}, which essentially involves finding ways of altering the input to a language model such that it matches templates that it has seen during pre-training.  Prompting a model correctly can give stronger performance in tasks,  help with controllability in the case of text generation, and is more parameter-efficient and data-efficient than fine-tuning \cite{li2021prefix}. One type of prompting is \textit{prefix prompting} \cite{li2021prefix,lester2021power}.  Prefix prompting consists of modifying a language model's input (i.e., prefix) by adding additional content. This can be explicit hard prompts (i.e., actual words such as "give a happy review") or soft prompts (i.e., embeddings that are input into a model and can be trained to converge on some virtual template or virtual prompt that can help the model). 


We introduce the idea of a \textit{hint}, a hybrid of hard and soft
prompts. We define a \textit{hint} as an additional input to a model in the form of a part(s) of an assertion that a
model has to predict, along with special identifier tokens for these parts, wrapped within parenthesis characters.  Syntactically, a \textit{hint} would take the form of: ``\textit{\textbf{(}[subject symbol, {subject}], [relation symbol, {relation}], [object symbol, {object}]\textbf{)}}" where the actual
content of the \textit{hint}, between the parenthesis, would be a
permutation of all but one of the elements of the target tuple during training. In the case of supplying hints to GLUCOSE, we include a special \textit{specific} or \textit{general} token which determines whether the part of the hint belongs to a story \textit{specific} assertion or a \textit{general} assertion.
For clarity, an example of a \textit{hint} for the hockey example using the GLUCOSE formulation would be ``\textit{(\textless{}$|$specific$|$\textgreater{}\textless{}$|$subj$|$\textgreater{}The\ red\
team\ scores, \textless{}$|$general$|$\textgreater{}\textless{}$|$obj$|$\textgreater{}People\_A\ win\ the\ game)}".  Altogether, the model input would be:
\begin{quote}
\tiny
\textbf{Model Input:} 1: The hockey game was tied up. The {red team} had the puck. They sprinted down the ice. They cracked a shot on goal! *\textit{They scored a final goal!}* \textit{(\textless{}$|$specific$|$\textgreater{}\textless{}$|$subj$|$\textgreater{}\textbf{The\ red\
team\ scores},  \textless$|$general$|$\textgreater{}\textless$|$obj$|$\textgreater{}\textbf{People\_A\ win\ the\ game})}\\\\
\textbf{Model Target/Output:} \textbf{The red team scores}, {Causes/Enables}, {The red team wins the game} ** {People\_A score}, {Causes/Enables}, \textbf{People\_A win the game}\\
\end{quote}
\vspace{-15pt}
\textit{Hints} are provided during training by sampling a
binomial distribution (with $p=0.5$) for each element in a minibatch, which
determines whether to give a \textit{hint} or not.  The actual content
of the \textit{hint} would then be generated by random sampling without replacement of up
to all but one of the elements in a target tuple. Once more, this only happens during training, later on one can supply whatever is desired as a hint to guide the generation of the commonsense inference. 

We hypothesize that this scheme of \textit{hinting} strikes a balance between the model recalling information from its pre-training, with information that it may not have seen that may only be present in the target tuple.  Additionally, by providing and fine-tuning a model on the combination of hard and soft prompts, a generative language model can be guided to ``talk" about a certain  {subject},  {object}, or  {relation}, thus enabling finer control of the models in downstream applications. We note that the approach was designed to be simple to implement and to give control when generating text. In the following sections, we give some background and follow this with a set of experiments to show the effects of \textit{hinting} for the ParaCOMET and GLUCOSE datasets, and finally, analyze the results, and present future directions for this work. 
Concretely, our contributions are:\begin{itemize}
    
    \item A hybrid prefix prompting technique called \textit{hinting} that provides a partial assertion to augment data for contextual commonsense inference, and

    \item Demonstrating that \textit{hinting} improves the performance for contextual commonsense inference as measured by automated metrics and is comparable in human-based metrics.
\end{itemize}

\section{Related Work}

\subsection{Prompting}
\label{sec:prompting}
Recently, there has been a shift in paradigm in Natural Language Processing from pre-training and fine-tuning a model, to pre-training, prompting, and predicting \cite{liu2021pre}.  One primary reason for this shift is the creation of ever-larger language models, which have become computationally expensive to fine-tune. Prompting can be described as converting a pre-trained language model input sequence into another sequence that resembles what the language model has seen during pre-training. Overall, most prompting research is focused on formulating the task as a \textit{cloze} (fill-in-the-blanks) task.  However, we consider the task of language generation, an open-ended formulation. 

Recall that prefix prompting modifies the input to a language model, by adding either a hard prompt (additional words to the input sequence)\cite{shin2020autoprompt} or a soft prompt (i.e., adding trainable vectors that represent, but are not equivalent to, additional words) \cite{li2021prefix,lester2021power,liu2021pre}. 

Unlike classic prefix prompting, \textit{hinting} uses both hard and soft prompts. The soft prompts are in the form of symbols that represent the different parts of the assertion (i.e.,  {subject},  {relation type}, and  {object}), and the hard prompts are in the form of the actual parts of the assertion that are selected to be appended as part of the \textit{hint}.  Our work is similar to KnowPrompt \cite{DBLP:journals/corr/abs-2104-07650}, except that they use a masked language model and soft prompts for relationship extraction. AutoPrompt \cite{shin2020autoprompt} is also similar, but finds a set of "trigger" words that give the best performance on a \textit{cloze}-related task, whereas we provide specific structured input for the model to guide text generation. We additionally note that although there are prompt-based relation extraction models \cite{chen2021adaprompt}, we are performing a different task which is contextual commonsense inference. 
Another recent contribution, P-Tuning \cite{liu2023gpt}, shares similarities with our approach by combining trainable continuous prompt embeddings with discrete prompts. Both hinting and P-tuning share the overarching objective of enhancing prompt learnability.
PTR, or prompt-tuning with rules \cite{han2021ptr}, shares similarities with hinting as it involves encoding prior knowledge about tasks and classes. Similarly to PTR, hinting also introduces additional information to provide the model with contextual understanding of the relationships between words.

\subsection{Controllable Generation}
Controllable generation can be described as ways to control a language model's text generation given some kind of guidance.  One work that tries to implement controllable generation is CTRL \cite{keskar2019ctrl}.  The authors supply control signals during pre-training of a general language model.  A body of work in controllable generation has focused on how it can be used for summarization.  Representative work that uses techniques similar to ours is GSum \cite{dou2020gsum}. 

In contrast to GSum, our method is model independent, allows for the source document to interact with the guidance signal, and contains soft prompts in the form of trainable embeddings that represent the parts of a tuple. The GSum system gives interesting insight into the fact that highlighted sentences, and the provision of triples, does in fact help with the factual correctness of abstractive summarization.  We make the distinction that \textit{hinting} falls more under prompting for the reason that we utilize additionally the trainable soft embeddings rather than purely additional hard tokens and that our task of contextual commonsense generation is not explored in the controllable generation works, whose main focus is on controlling unstructured text generation. Some works that are in this area are also \cite{peng2018towards} who utilize what they call "control factors" as keywords or phrases that are supplied by a human-in-the-loop to guide a conversation.
Another work, Diffusion-LM \cite{NEURIPS2022_1be5bc25}, developed a language model that utilizes a sequence of Gaussian noise vectors that gradually transform into words, creating an organized structure to guide what the model generates.
More similar to our work, but tailored for the task of interactive story generation and without trainable soft-embeddings, is the work by \cite{brahman-etal-2020-cue} which uses automatically extracted keywords to generate a story. Future work we could possibly utilize the automatic keyword extraction to supply parts of a hint, rather than our approach of complete parts of an assertion, and expand this to utilize synonyms of keywords. Lastly, there is the work by  \cite{see-etal-2019-makes} which looks at controllable text generation for the purpose of conversation and utilizes an embedding give quantitative control signals as part of conditional training.

\subsection{Discourse-aware/Contextual commonsense inference}
Commonsense inference is the task of generating a commonsense assertion.  Discourse-aware/contextual commonsense inference is the task of, given a certain narrative or discourse, inferring commonsense assertions that are coherent within the narrative\cite{Gabriel_PC}. This task is particularly hard because commonsense knowledge may not be explicitly stated in text \cite{liu2004conceptnet} and the model needs to keep track of entities and their states either explicitly or implicitly.  Research into the knowledge that pre-trained language models learn has yielded good results in that they do contain various types of factual knowledge, as well as some commonsense knowledge\cite{da2019cracking,petroni2019language,davison2019commonsense}. The amount of commonsense knowledge in these models can be improved by supplementing sparsely covered subject areas with structured knowledge sources such as ConceptNet \cite{speer2017conceptnet,davison2019commonsense}. 

Knowing that these pre-trained language models may contain some commonsense information has led to the development of knowledge models such as COMET\cite{bosselut2019comet}.  This line of research has been extended from the  sentence-by-sentence level in COMET, to the paragraph-level in ParaCOMET \cite{Gabriel_PC}. Contemporaneously, GLUCOSE \citet{mostafazadeh-etal-2020-glucose} builds a dataset of commonsense assertions that are contextualized to a set of stories, and generalized. More recently, the idea of  knowledge models \cite{da2021analyzing} or models that can be leveraged to generate commonsense assertions has been gaining track. One recent approach has been kogito \cite{ismayilzada2022kogito}. Kogito is a toolkit for commonsense inference, which permits training and access of  similar to COMET, along with providing tools for selecting a subject and a relation. Kogito utilizes the same formulation as the ParaCOMET work, in which a subject and a relation are provided. However, kogito does not tackle the more complicated general commonsense inference as in GLUCOSE. With our work, we could provide kogito with a framework to be able to train models that can perform this type of controllable, generalized inference and improve the overall training.


\section{Modeling}
\begin{table*}[]
\captionsetup{font=tiny}

\resizebox{\textwidth}{!}{\begin{tabular}{|l|l|}
\hline
Model Input                                                                                                                                                                                                                                                                                                                                                      & Model Output                                                                                                                                                                                                                                                                     \\ \hline
\begin{tabular}[c]{@{}l@{}}7: The hockey game was tied up.  \\ The red team had the puck. They sprinted down the ice. \\ They cracked a shot on goal!. * They scored a final goal!. *\end{tabular}                                                                                                                        & \begin{tabular}[c]{@{}l@{}}They scored a final goal \textgreater{}Causes\textgreater They feel(s) happy ** \\ Some People\_A scored a final goal \textgreater{}Causes\textgreater Some People\_A feel(s) happy\end{tabular}                                                      \\ \hline
\begin{tabular}[c]{@{}l@{}}7: The hockey game was tied up.  \\ The red team had the puck. They sprinted down the ice. \\ They cracked a shot on goal!. * They scored a final goal!. *\\ \textbf{hint: (\textless{}$|$specific$|$\textgreater{}\textless{}$|$subj$|$\textgreater {the red team scores the final goal})}\end{tabular}          & \begin{tabular}[c]{@{}l@{}}{\textbf{the red team scores the final goal}} \textgreater{}Causes\textgreater the red team feel(s) happy ** \\ Some People\_A (who are a team) score the final goal \textgreater{}Causes\textgreater Some People\_A feel(s) happy\end{tabular}                  \\ \hline
\begin{tabular}[c]{@{}l@{}}7: The hockey game was tied up.  \\ The red team had the puck. They sprinted down the ice. \\ They cracked a shot on goal!. * They scored a final goal!. *\\ \textbf{hint: (\textless{}$|$general$|$\textgreater{}\textless{}$|$subj$|$\textgreater{\textbf{ Something\_A (that is a point)}})}\end{tabular}               & \begin{tabular}[c]{@{}l@{}}They scored a final goal \textgreater{}Causes\textgreater They feel(s) happy ** \\ {\textbf{Something\_A (that is a point))} is scored} \textgreater{}Causes\textgreater Some People\_A feel(s) happy\end{tabular}                                               \\ \hline
\end{tabular}}
\caption{Example of inputs and outputs for the GLUCOSE trained model with hints. The \textit{hint} is \textbf{bolded}). Without a hint we can see that the model does a good faith attempt with the given context and with a "hint" it is guided.}
\label{tab:illustrativeexample}
\end{table*}


\subsection{Task}
We now detail the task of Contextual Commonsense Inference.  We are given a story $S$ composed of $n$ sentences, $S=\{S_1,S_2,…,S_n\}$ , a target sentence from that story, $S_t$, where $S_t \in S$, and a dimension/relation type $R$. Given all this, we want to generate a tuple in the form of $({subject}, {R},{object})$ that represents an assertion, present or implied, in $S_t$ given the context $S$, and the relation type $R$. 

We run tests with two variations of this task, one is the ParaCOMET variation and the other the GLUCOSE variation. In the ParaCOMET experiments, we represent $S_t$ with a special token.  Additionally, we only generate the {$object$} of the tuple. In our GLUCOSE experiments, we represent $S_t$ by marking it with $*$ on the left and right of the actual target sentence. Additionally, we generate at most two \textit{{subject}, {R}, {object}} tuples: one that is the context-specific tuple, and the other is the general tuple, separated by two asterisks (**). An example of how the inputs look for both datasets can be seen earlier in the Introduction. 

\subsection{Hinting}
\label{sec:hinting1}

The mechanism we present in this work, called \textit{hinting}, is a kind of mixed/hybrid prompting for generative language models.  Prompting is essentially supplying additional text (i.e. prompts) to a language model to aid/guide it in a specific task.  In our case, we opt to give a ``hint", as to what the assertion that we want to predict contains, at the end of our input text.  We chose placing the hint at the end of the input for simplicity in dataset processing, but it can be placed anywhere and we leave it as future work to explore the effects of placing hints possibly next to the target sentence or at the beginning of the input. Hinting can be seen as a hybrid of prompting the generative model with hard prompts composed of parts of what should be predicted along with soft prompts of symbols that represent those parts. These symbols are for the \textit{subject}, \textit{relation (R)}, and \textit{object} respectively. These soft prompts utilize untrained embeddings for the task. We structure hinting this way such that, after training, whenever a \textit{hint} is given, the model can be guided to generate assertions about the \textit{hint's} content based on the target sentence and context.

To balance the model's reliance on the context, its knowledge, and the \textit{hint}, we determine whether to supply the \textit{hint} by sampling a binomial distribution ($p=0.5$). Thus, we can control the frequency of when to supply a \textit{hint}. Additionally, the content of the \textit{hint} is determined by random sampling of permutations of components, up to a maximum of all but one component. Since our task is to predict the tuple, we do not want to make the model overly reliant on \textit{hints} for the answer.  

\subsection{An example of Hinting}
\label{sec:hinting}
A simple example of \textit{hinting} is the following:
\begin{quote}
    \tiny
    \textbf{Story:} \textit{The hockey game was tied up. The red team had the puck. They sprinted down the ice. They cracked a shot on goal! They scored a final goal!} 
    
    \textbf{Target sentence:} \textit{They scored a final goal!}

    \textbf{Target assertion:} \textit{(subject: {the red team}, relation: {are capable of}, object: {winning the game}.)}
\end{quote}

A \textit{hint} can be any permutation of the target assertion, except the complete assertion, along with some symbol that indicates which part it is:
\begin{quote}
    \tiny

\textbf{Possible Hints:}\\\textit{(\textless{}$|$subj$|$\textgreater{} {the\ red\ team}),\\ 
(\textless{}$|$obj$|$\textgreater{} {winning\ the\ game}),\\ (\textless{}$|$rel$|$\textgreater{} {capable\ of}),\\
(\textless{}$|$subj$|$\textgreater{} {the\ red\ team},  \textless{}$|$rel$|$\textgreater{} {capable\ of}),\\ (\textless{}$|$subj$|$\textgreater{} {the\ red\ team}, \textless{}$|$obj$|$\textgreater{} {winning\ the\ game}),\\ (\textless{}$|$rel$|$\textgreater{} {capable\ of}, \textless{}$|$obj$|$\textgreater{} {winning\ the\ game})}
\end{quote}

Sampling randomly from one of those permutations, a \textit{hint} for the given story, target sentence and target assertion, yields the following:
\begin{quote}
    \tiny
    \textbf{Hint:} \textit{(\textless{}$|$subj$|$\textgreater{} {the\ red\ team},  \textless{}$|$rel$|$\textgreater{} {capable\ of})}

\end{quote}

Putting everything together, the input for the model would be: 
\begin{quote}
    \vspace{-1.75pt}
    \tiny

    \textbf{Story with Hint:} \textit{The hockey game was tied up. The red team had the puck. They sprinted down the ice. They cracked a shot on goal! They scored a final goal! \textbf{(\textless{}$|$subj$|$\textgreater{} {the\ red\ team},  \textless{}$|$rel$|$\textgreater{} {capable\ of})}}. 
\end{quote}

We note that this is a general version of how the hinting mechanism works. The dataset specific hints that we utilize are described in the Appendix. 

\subsection{Hinting with Synonyms and Antonyms}
 To explore other ways of hinting, we devised another technique by swapping parts of the hint with synonyms and antonyms. We used the synsets and antsets from the WordNet \cite{fellbaum2010wordnet} knowledge base to find synonyms and antonyms respectively. In the case that there is no viable synonym or antonym we keep the original word. To introduce synonyms and antonyms, the process is as follows. When a hint is generated, we copy it and we decide whether to supply it with synonyms and/or antonyms by swapping the words in the hint prompt with their synonyms or antonyms by sampling from a binomial distribution (we explored various values for $p$, with 0.5 being the most effective) to control the frequency of hints with synonyms/antonyms. The reasoning for this is that if we only supply synonyms/antonyms, we lose the control that hinting provides, so we introduce synonyms and antonyms at a certain rate to have the model still be controllable, while seeing “new” information through the synonyms/antonyms. After the swapping process, we insert a soft prompt, \textit{\textless{}$|$syn$|$\textgreater{}} or \textit{\textless{}$|$ant$|$\textgreater{}}, at the end of the substituted prompt to signal that the model is using a synonym or an antonym respectively. We give an example of how these hints look in the Appendix.

\subsection{Models}
For our first set of experiments, we utilize the ParaCOMET \cite{Gabriel_PC} dataset and the framework with the same GPT-2 model as ParaCOMET, along with a T5 \cite{2020t5} model to observe the effects of \textit{hinting} in a sequence-to-sequence formulation of the dataset. We use the off-the-shelf (Huggingface \cite{wolf2019huggingface}) pre-trained ``base" version of these models for efficiency. For our second set of experiments with the GLUCOSE dataset, we also used the T5 model, as was done in GLUCOSE.

\section{Experimental Setup}

To show the effectiveness of \textit{hinting} we use the following setups. First we utilize the original ParaCOMET dataset and setup and adding \textit{hints with/without synonyms and antonyms} . The ParaCOMET setup consists of given a story $S$ composed of $n$ sentences, $S=\{S_1,S_2,…,S_n\}$, a relation type $R$, and a target sentence token (i.e. \textit{\textless{}$|$sent0$|$\textgreater{}}, \textit{\textless{}$|$sent1$|$\textgreater{}}, …, \textit{\textless{}$|$sent(n-1)$|$\textgreater{}}). In the ParaCOMET dataset, we must predict the \textit{object} of a triple, utilizing implicitly the sentence as a  {subject} and explicitly the supplied sentence symbol and  {relation} $R$ symbol. 

Within this framework, after the relation $R$, we add our \textit{hint} between parenthesis (i.e. “([\textit{hint}])”).  In this framing, our \textit{hint} can be composed of: a \textit{subject} symbol (\textless{}$|$subj$|$\textgreater{}) along with the target sentence to serve as a \textit{subject}, a relation symbol (\textless{}$|$rel$|$\textgreater{}) along with the \textit{relation}  $R$, or an \textit{object} symbol (\textless{}$|$obj$|$\textgreater{}) along with the \textit{object} of the triple. Using the hockey example a possible \textit{hint} in this set of experiments would be: "({\textless{}$|$rel$|$\textgreater{} \textless{}$|$xEffect$|$\textgreater{}},{\textless{}$|$obj$|$\textgreater{} they win the game})". In the case that we add a synonym and/or antonym, we do the appropriate replacement and add the \textit{\textless{}$|$syn$|$\textgreater{}} or \textit{\textless{}$|$ant$|$\textgreater{}} tags. It is possible, although we leave for future work, to include both the hint \textit{and} the synonym/antonym hint as part of the prompt.

\begin{table*}[ht]
\tiny

\captionsetup{font=tiny}

\resizebox{\textwidth}{!}{
\begin{tabular}{|l|c|c|c|c|c|c|c|c|}
\hline
\textbf{Model}       & \textbf{BLEU1}   & \textbf{BLEU2}   & \textbf{BLEU}    & \textbf{METEOR}  & \textbf{ROUGE-1} & \textbf{ROUGE-2} & \textbf{ROUGE-L} & \textbf{ROUGE-LSUM} \\ \hline
GPT-2                & 90.507           & 83.464           & 42.406           & 68.858           & 61.985           & 51.051           & 61.951           & 61.952              \\ \hline
GPT-2 Hint           & 91.410*          & 85.934*          & 43.519*          & 69.558*          & 63.501*          & 52.763*          & 63.481*          & 63.477*             \\ \hline
GPT-2 Hint+Synonyms  & \textbf{91.782*} & \textbf{86.610*} & 43.611*          & 69.592*          & 63.559*          & 52.825*          & 63.546*          & 63.539*             \\ \hline
GPT-2 Hint+Antonyms  & 91.622*          & 86.491*          & \textbf{43.652*} & \textbf{69.605*} & \textbf{63.636*} & \textbf{52.915*} & \textbf{63.619*} & \textbf{63.619*}    \\ \hline
GPT-2 Hint+Syn.+Ant. & 91.497*          & 85.850*          & 43.495*          & 69.428*          & 63.311*          & 52.579*          & 63.292*          & 63.290*             \\ \hline \hline
T5                   & 94.133           & 80.462           & 38.585           & 65.815           & 56.095           & 44.636           & 56.078           & 56.079              \\ \hline
T5 Hint              & 93.851*          & 81.284           & 39.622           & 66.409*          & 57.623           & 46.441           & 57.591           & 57.593              \\ \hline
T5 Hint+Synonyms     & 94.562           & \textbf{83.953}  & \textbf{40.656*} & \textbf{67.381*} & \textbf{58.385*} & \textbf{47.565*} & \textbf{58.348*} & \textbf{58.350*}    \\ \hline
T5 Hint+Antonyms     & \textbf{95.482*} & 82.286*          & 38.735           & 65.917           & 56.223           & 44.748           & 56.191           & 56.190              \\ \hline
T5 Hint+Syn.+Ant.    & 94.543           & 81.193           & 39.153           & 66.465           & 56.805           & 45.588           & 56.765           & 56.766              \\ \hline
\end{tabular}
}
\caption{Averages of 4 different seeds over 5 epochs for the ParaCOMET dataset from \cite{Gabriel_PC} for different hinting configurations.  The largest scores are \textbf{bolded} and scores that are significantly different from the non-hinted model have an asterisk (*) next to them. 
We can see from the results that \textit{hinted} systems tend to achieve higher performance even if slightly and in some cases significantly. We also see that hinting training with synonyms tends to be more beneficial than with antonyms, but both improve the performance of hinting.}
\label{tab:expset1}
\end{table*}

%
\begin{table*}[]

\tiny \centering

\captionsetup{font=tiny}
\begin{tabular}{|l|c|c|c|c|c|c|}
\hline
Model             & BLEU             & METEOR          & ROUGE-1         & ROUGE-2         & ROUGE-L         & ROUGE-LSUM      \\ \hline
T5                & 75.998           & 80.683          & 80.282          & 67.349          & 78.624          & 78.622          \\ \hline
T5 Hint           & 76.266           & \textbf{80.971} & \textbf{80.543} & \textbf{67.725} & \textbf{78.894} & \textbf{78.908} \\ \hline
T5 Hint+Synonyms  & 75.848           & 80.879          & 80.511          & 67.605          & 78.844          & 78.843          \\ \hline
T5 Hint+Antonyms  & \textbf{76.282*} & 80.840*         & 80.448          & 67.720          & 78.846*         & 78.858*         \\ \hline
T5 Hint+Syn.+Ant. & 76.138           & 80.798          & 80.365          & 67.607          & 78.746          & 78.756          \\ \hline
\end{tabular}

\caption{Averages of 4 different seeds over 5 epochs for the GLUCOSE dataset from \cite{mostafazadeh-etal-2020-glucose} for different hinting configurations.  The largest scores are \textbf{bolded} and scores that are significantly different (p\textless{}0.05 in T test) from the non-hinted model have an asterisk (*) next to them. Once more we see that hinted models tend to perform better or comparable to not hinting. Here we see that the benefits of adding synonyms and antonyms to training are not as pronounced as in Table \ref{tab:expset1}.}
\label{tab:expset2}
\end{table*}

\begin{table*}[]
\tiny
\captionsetup{font=tiny}

\resizebox{\textwidth}{!}{

\begin{tabular}{|l|l|l|l|l|l|l|}
\hline
\textbf{Model}             & \textbf{General Subject} & \textbf{General Subj.+Rel.} & \textbf{Specific Subject} & \textbf{Specific Subj.+Rel.} & \textbf{\begin{tabular}[c]{@{}l@{}}Specific Subj.+Rel. \& \\ General Subj.\end{tabular}} & \textbf{\begin{tabular}[c]{@{}l@{}}Specific Subj.+Rel. \& \\ General Subj.+Rel.\end{tabular}} \\ \hline
T5                & 51.047                   & 56.256                      & 58.386                    & 67.287                       & 59.345                                                                                   & 61.555                                                                                        \\ \hline
T5 Hint           & 99.928                   & \textbf{99.974}             & 99.981                    & \textbf{100.000}             & 99.963                                                                                   & 99.977                                                                                        \\ \hline
T5 Hint+Synonyms   & 99.892                   & 99.931                      & \textbf{100.000}          & \textbf{100.000}             & 99.689                                                                                   & 99.987                                                                                        \\ \hline

T5 Hint+Antonyms   & 99.939                   & 99.839                      & \textbf{100.000}          & \textbf{100.000}             & \textbf{99.988}                                                                          & 99.987                                                                                        \\ \hline
T5 Hint+Syn.+Ant. & \textbf{99.988}          & 99.854                      & 99.973                    & \textbf{100.000}             & 99.978                                                                                   & \textbf{99.990}                                                                               \\ \hline
\end{tabular}
}
\caption{Results of control tests to see if elements provided in hint show up in the output. All results are in BLEU. We see that models that are given hints during the training retain controllability in evaluation. We see that if a certain element of a relation is given in the hint, the model does a good effort, even with synonym and antonym training, of utilizing that in its output.}
\label{tab:expset3}

\end{table*}

\begin{table*}
\captionsetup{font=tiny}

\tiny
\centering
\begin{tabular}{|l|c|c|}
\hline
\textbf{Model}    & \textbf{Avg. Rating} & \textbf{Plausibility (\textgreater{}=3)} \\ \hline
T5                & 3.77                 & 98.66\%                                  \\ \hline
T5 Hint           & \textbf{3.90}                 & 96.05\%                                  \\ \hline
T5 Hint+Synonym   & 3.85                 & 94.73\%                                  \\ \hline
T5 Hint+Antonym   & 3.76                 & 97.36\%                                  \\ \hline
T5 Hint+Syn.+Ant. & 3.86                 & \textbf{100.00}\%                                 \\ \hline
\end{tabular}
\caption{Results of human evaluation of GLUCOSE dataset model. The largest scores are \textbf{bolded}. We sampled 75 test points for each model from the test dataset and had them infer assertions. Humans judged these assertions on a 5 point Likert scale where above 3 was plausible similar to  \cite{Gabriel_PC}. We had 3 raters on each inference and utilized the median of the rating.  We can see that hinting provides similar plausibility and tends to be rated higher than not hinting at all.}
\label{tab:expset4}

\end{table*}
In our GPT-2 experiments, we utilize the same cross-entropy loss as in \cite{Gabriel_PC}.  We note that we also utilize a sequence-to-sequence \cite{sutskever2014sequence} formulation for the T5 model. This in contrast to the GPT-2-based system requires encoding a source sequence (i.e., story, target sentence, and relation symbol), and decoding it into a target sequence (i.e., the \textit{object} of an assertion). For the T5 model, we add the prefix "source:” before the story $S$, and the prefix "hint:" for placing our \textit{hints}. In addition, whenever there is a synonym or antonym added, it is added as another prefix (``synonym:[synonym]" or ``antonym:[antonym]"). For simplicity, we construct the same "heuristic" dataset as ParaCOMET which utilizes a heuristic matching technique to align ATOMIC \cite{sap2019atomic} triples to story sentences. 

Secondly, we utilize the formulation utilized in GLUCOSE \cite{mostafazadeh-etal-2020-glucose}.  The formulation utilizes the T5 model in a sequence-to-sequence formulation once more.  In this formulation, the source text is composed of a prefix of a dimension to predict $D \in {1,2,…10}$\footnote{The definition for these numbers is in \cite{mostafazadeh-etal-2020-glucose}}, followed by the story $S$ with the marked target sentence. The target sentence, $S_t$, is marked with ``$*$" before and after the sentence. An example input is: ``1: The first sentence. \textit{*The target sentence. *} The third sentence.". This task is slightly different from the ParaCOMET one, in that in addition to predicting a context \textit{specific} triple, the model has to predict a \textit{generalized} triple. 

In the GLUCOSE task we have to infer both \textit{general} and context \textit{specific}  \textit{subject},  \textit{object} and \textit{relation} elements. For our \textit{hints} we provide up to five out of these six elements while training, along with a symbol that represents whether it is the \textit{subject}, \textit{object} or a  \textit{relation}, and another symbol that represents whether it is part of the \textit{general} or \textit{specific} assertion. We add our \textit{hint} after the story $S$, utilizing the prefix ``hint:" and supplying the \textit{hint} between parenthesis. Given the hockey story, an example of a \textit{hint} for GLUCOSE can be: ``\textbf{\textit{(\textless{}$|$general$|$\textgreater{} \textless{}$|$obj$|$\textgreater{} {People\_A win a Something\_A})}}". Hyperparameter configuration details can be seen in the Appendix.

Thirdly, for testing the controllability of the model, we train 5 models on the GLUCOSE data: one without hints, one with hints, one with hints and synonyms, one with hints and antonyms, and one with hints, synonyms, and antonyms. To test the control, we make a synthetic test in which we utilize the GLUCOSE testing data, and supply the model with hints for the specific subject and/or relation and/or the general subject and/or relation. The test measures the overlap between the elements provided in the hint and the elements present in the output. We note that this is a synthetic benchmark which demonstrates that the model is capable of incorporating the hint into its output accordingly.

Lastly, we ran a tiny Mechanical Turk study similar to the one presented in the original ParaCOMET \cite{Gabriel_PC} in which a human judges the plausibility of the generated assertion based on the context on a 5-point Likert scale: obviously true (5), generally true (4), plausible (3), neutral or unclear (2), and doesn’t make sense (1). We present the results in the same manner where Table \ref{tab:expset3} displays the percent of inferences judged as plausible or true (3-5), and the average rating per inference. Participants were given \$0.1  to complete the task. We give an image of the HIT in the Appendix.  We sample from the GLUCOSE test set, 75 entries randomly. Then, for all the trained models, we generate assertions for these entries and hand them off to 3 human raters. 
\section{Results and Analysis}
\subsection{Experiment 1: ParaCOMET with \textit{hints}}
The aggregated results for this set of experiments can be found in Table \ref{tab:expset1}. We can see here that on average, \textit{hinting} does tend to improve the score even if slightly.  It seems that providing a \textit{hint} is beneficial and not detrimental for contextual commonsense inference.   Given the way that this task is framed, a possibility that could explain the relative similarity of the performances, is that \textit{hinting} in this formulation \textbf{only} adds the \textit{object} of the triple as additional possible data that the model may see during training; the  {subject} and the  {relation} can be repeated with \textit{hinting}.   We note that the performance of the T5 model was less, and we believe that it may be lack of hyperparameter tuning, as it was seen that the model was sensitive to the learning rate and had to use a higher than usual learning rate. In these tests, we also note that hinting with synonyms tends to consistently improve the performance even further than just plain hinting, indicating that the model benefits from making associations of related concepts. We see that hinting with antonyms also tends to be beneficial, but the benefit is not as consistent as with synonyms. Interestingly, we see that combining both synonym and antonym training does not bring the best of both worlds, but more closely an average between the performance of hinting with either. 

\subsection{Experiment 2: GLUCOSE with \textit{hints}}
The aggregated results for this set of experiments can be found in Table \ref{tab:expset2}. Once more we notice that \textit{hinting} (with and without synonyms and/or antonyms) does tend to improve the performance of the contextual commonsense inference task.  This suggests that \textit{hinting} is indeed beneficial for the task of contextualized commonsense inference, especially when faced with the harder task of generating both a general and context dependent assertion.  We believe that this improvement is because \textit{hinting} gives the model the clues it may need to decide on what to focus or attend to, to generate useful inferences. 

\subsection{Experiment 3: Controllability}
In Table \ref{tab:expset3} we see the results of our synthetic controllability test. We see that the model without hinting tends to predict about relevant things (indicated by the BLEU scores above 50), however when hints are injected, the model always predicts about what the hint was suggesting (indicated by the nearly perfect scores). We also see that supplying synonyms and antonyms does not decrease the controllability of the models. 
\subsection{Experiment 4: Human Judgements}
The results for a tiny Mechanical Turk study for human evaluation of model inferences can be seen in Table \ref{tab:expset4}.  Overall we can see here that hinted systems are judged as slightly higher in plausibility. We also see upon looking some of the inferences that the hinted model tends to be more general and provide shorter  responses than the non-hinted model (e.g., hinted inference: ``satisfied" vs. non-hinted inference: ``happy and satisfied"). 

\section{Discussion}
\subsection{Why \textit{hint}?}
From the results of our experiments, we can see that \textit{hinting} tends to increase the performance of contextualized commonsense inference at least with regards to automated metrics and does not significantly degrade or improve human judgements.  This brings the question of: Why \textit{hint} at all? The primary reason is for controllability in the generation. By supplying these \textit{hints}, we are teaching the model pay attention and generate inferences about a certain {subject}, {relation}, or  {object}. This in turn, after training, can be leveraged by a user or downstream application to guide the model to generate assertions from parts that are manually supplied. Although this is not very clear within the ParaCOMET formulation, it becomes clearer in the GLUCOSE formulation of the problem.  We give an illustrative example of the usefulness of \textit{hinting} in Table \ref{tab:illustrativeexample}. We can see that by giving a model the \textit{hint}, the model could be capable of inferring about information that may not be present in the story. We note that this behavior is useful in downstream tasks such as story understanding and contextual knowledge graph generation in which we may need a model to have a specific {subject} or {object} . Lastly, \textit{hinting} was designed to be simple to implement, and is model independent.

\subsection{Is  \textit{hinting} optimal?}

This work was a proof of concept for this technique.  We acknowledge there is a large body of research on the area of prompting. The way the \textit{hinting} mechanism was designed however, leaves much space to explore alternate mechanisms such as AutoPrompt\cite{shin2020autoprompt},  including additional soft prompts such as those in \citet{li2021prefix}, or even replacing the contents of the hint with synonyms or related words. Because of the naivety of the approach, we do not think it is an optimal approach, and there is a large body of research that points to manual templating of prompts being less effective than learned prompts \cite{liu2021pre}. However, from our tests, our approach does not degrade performance, and only improves it.

\section{Conclusion}

In this work we presented \textit{hinting}, a simple hybrid prompting mechanism that consists of appending parts of a target tuple into an input sequence for the task of contextual commonsense inference. We showed that \textit{hinting} tends to improve performance in automated metrics and provides comparable performance with human-based judgements. With this, we open the doors for exploring prompting within the realm of contextual commonsense inference. 

The hinting system design acknowledges areas for improvement, particularly in developing smarter strategies for selecting when and what to hint, and enhancing the hint with additional soft prompts. Future work and exploration is further described in the Appendix \ref{appendix:futurework}.
\section{Ethics Statement}
In this work, we propose a mechanism called ``hinting" to create a controllable contextual commonsense inference model. Our goal is to improve the usability of contextual commonsense inference models in downstream applications. However, it's important to note that our mechanism may be subject to limitations due to biases existing in the knowledge bases used (e.g., ATOMIC, and GLUCOSE).

While our model is controllable to some extent, it could potentially generate harmful or incorrect assertions as a result of the conditioned biases. Additionally, since our model generates text, it might produce erroneous statements. We did not analyze the degree of these biases or incorrect inferences in this study; however, we utilize well-vetted knowledge bases which should minimize any significant negative impacts on performance.
\section{Limitations}
We note that our work has some limitations. One of these is the length of the stories that were given for the task of contextual commonsense inference. Many of these stories are around 5 sentences long. This may in turn harm the generalization of the effectiveness of this technique to longer stories. We also note that this technique was designed for language models that are not extremely large (>7B parameters) which have recently shown their effectiveness in a wide variety of tasks, however, since these are prompts, the hinting technique can be utilized as part of example prompts for these extremely large models. Lastly, one limitation of this method is that the inferences that are produced have no way of being evaluated on their relevance and truthfulness. This is to be addressed in future work with a classifier for these assertions. 


\bibliography{main}

\begin{thebibliography}{39}
\expandafter\ifx\csname natexlab\endcsname\relax\def\natexlab#1{#1}\fi

\bibitem[{Banerjee and Lavie(2005)}]{banarjee2005}
Satanjeev Banerjee and Alon Lavie. 2005.
\newblock \href {https://www.aclweb.org/anthology/W05-0909} {{METEOR}: An automatic metric for {MT} evaluation with improved correlation with human judgments}.
\newblock In \emph{Proceedings of the {ACL} Workshop on Intrinsic and Extrinsic Evaluation Measures for Machine Translation and/or Summarization}, pages 65--72, Ann Arbor, Michigan. Association for Computational Linguistics.

\bibitem[{Bosselut et~al.(2019)Bosselut, Rashkin, Sap, Malaviya, Çelikyilmaz, and Choi}]{bosselut2019comet}
Antoine Bosselut, Hannah Rashkin, Maarten Sap, Chaitanya Malaviya, Asli Çelikyilmaz, and Yejin Choi. 2019.
\newblock Comet: Commonsense transformers for automatic knowledge graph construction.
\newblock In \emph{Proceedings of the 57th Annual Meeting of the Association for Computational Linguistics (ACL)}.

\bibitem[{Brahman et~al.(2020)Brahman, Petrusca, and Chaturvedi}]{brahman-etal-2020-cue}
Faeze Brahman, Alexandru Petrusca, and Snigdha Chaturvedi. 2020.
\newblock \href {https://aclanthology.org/2020.aacl-main.59} {Cue me in: Content-inducing approaches to interactive story generation}.
\newblock In \emph{Proceedings of the 1st Conference of the Asia-Pacific Chapter of the Association for Computational Linguistics and the 10th International Joint Conference on Natural Language Processing}, pages 588--597, Suzhou, China. Association for Computational Linguistics.

\bibitem[{Chen et~al.(2021{\natexlab{a}})Chen, Xie, Zhang, Yan, Deng, Tan, Huang, Si, and Chen}]{DBLP:journals/corr/abs-2104-07650}
Xiang Chen, Xin Xie, Ningyu Zhang, Jiahuan Yan, Shumin Deng, Chuanqi Tan, Fei Huang, Luo Si, and Huajun Chen. 2021{\natexlab{a}}.
\newblock \href {http://arxiv.org/abs/2104.07650} {Adaprompt: Adaptive prompt-based finetuning for relation extraction}.
\newblock \emph{CoRR}, abs/2104.07650.

\bibitem[{Chen et~al.(2021{\natexlab{b}})Chen, Xie, Zhang, Yan, Deng, Tan, Huang, Si, and Chen}]{chen2021adaprompt}
Xiang Chen, Xin Xie, Ningyu Zhang, Jiahuan Yan, Shumin Deng, Chuanqi Tan, Fei Huang, Luo Si, and Huajun Chen. 2021{\natexlab{b}}.
\newblock Adaprompt: Adaptive prompt-based finetuning for relation extraction.
\newblock \emph{arXiv preprint arXiv:2104.07650}.

\bibitem[{Da et~al.(2021)Da, Bras, Lu, Choi, and Bosselut}]{da2021analyzing}
Jeff Da, Ronan~Le Bras, Ximing Lu, Yejin Choi, and Antoine Bosselut. 2021.
\newblock Analyzing commonsense emergence in few-shot knowledge models.
\newblock \emph{arXiv preprint arXiv:2101.00297}.

\bibitem[{Da and Kasai(2019)}]{da2019cracking}
Jeff Da and Jungo Kasai. 2019.
\newblock \href {https://doi.org/10.18653/v1/D19-6001} {Cracking the contextual commonsense code: Understanding commonsense reasoning aptitude of deep contextual representations}.
\newblock In \emph{Proceedings of the First Workshop on Commonsense Inference in Natural Language Processing}, pages 1--12, Hong Kong, China. Association for Computational Linguistics.

\bibitem[{Davison et~al.(2019)Davison, Feldman, and Rush}]{davison2019commonsense}
Joe Davison, Joshua Feldman, and Alexander~M Rush. 2019.
\newblock Commonsense knowledge mining from pretrained models.
\newblock In \emph{Proceedings of the 2019 Conference on Empirical Methods in Natural Language Processing and the 9th International Joint Conference on Natural Language Processing (EMNLP-IJCNLP)}, pages 1173--1178.

\bibitem[{Devlin et~al.(2019)Devlin, Chang, Lee, and Toutanova}]{devlin-etal-2019-bert}
Jacob Devlin, Ming-Wei Chang, Kenton Lee, and Kristina Toutanova. 2019.
\newblock \href {https://doi.org/10.18653/v1/N19-1423} {{BERT}: Pre-training of deep bidirectional transformers for language understanding}.
\newblock In \emph{Proceedings of the 2019 Conference of the North {A}merican Chapter of the Association for Computational Linguistics: Human Language Technologies, Volume 1 (Long and Short Papers)}, pages 4171--4186, Minneapolis, Minnesota. Association for Computational Linguistics.

\bibitem[{Dou et~al.(2021)Dou, Liu, Hayashi, Jiang, and Neubig}]{dou2020gsum}
Zi-Yi Dou, Pengfei Liu, Hiroaki Hayashi, Zhengbao Jiang, and Graham Neubig. 2021.
\newblock \href {https://doi.org/10.18653/v1/2021.naacl-main.384} {{GS}um: A general framework for guided neural abstractive summarization}.
\newblock In \emph{Proceedings of the 2021 Conference of the North American Chapter of the Association for Computational Linguistics: Human Language Technologies}, pages 4830--4842, Online. Association for Computational Linguistics.

\bibitem[{Fellbaum(2010)}]{fellbaum2010wordnet}
Christiane Fellbaum. 2010.
\newblock Wordnet.
\newblock In \emph{Theory and applications of ontology: computer applications}, pages 231--243. Springer.

\bibitem[{Gabriel et~al.(2021)Gabriel, Bhagavatula, Shwartz, Le~Bras, Forbes, and Choi}]{Gabriel_PC}
Saadia Gabriel, Chandra Bhagavatula, Vered Shwartz, Ronan Le~Bras, Maxwell Forbes, and Yejin Choi. 2021.
\newblock \href {https://ojs.aaai.org/index.php/AAAI/article/view/17521} {Paragraph-level commonsense transformers with recurrent memory}.
\newblock \emph{Proceedings of the AAAI Conference on Artificial Intelligence}, 35(14):12857--12865.

\bibitem[{Han et~al.(2021)Han, Zhao, Ding, Liu, and Sun}]{han2021ptr}
Xu~Han, Weilin Zhao, Ning Ding, Zhiyuan Liu, and Maosong Sun. 2021.
\newblock \href {http://arxiv.org/abs/2105.11259} {Ptr: Prompt tuning with rules for text classification}.

\bibitem[{Ismayilzada and Bosselut(2022)}]{ismayilzada2022kogito}
Mete Ismayilzada and Antoine Bosselut. 2022.
\newblock kogito: A commonsense knowledge inference toolkit.
\newblock \emph{arXiv preprint arXiv:2211.08451}.

\bibitem[{Keskar et~al.(2019)Keskar, McCann, Varshney, Xiong, and Socher}]{keskar2019ctrl}
Nitish~Shirish Keskar, Bryan McCann, Lav~R Varshney, Caiming Xiong, and Richard Socher. 2019.
\newblock Ctrl: A conditional transformer language model for controllable generation.
\newblock \emph{arXiv preprint arXiv:1909.05858}.

\bibitem[{Kingma and Ba(2015)}]{kingma2014adam}
Diederik~P. Kingma and Jimmy Ba. 2015.
\newblock Adam: A method for stochastic optimization.
\newblock \emph{CoRR}, abs/1412.6980.

\bibitem[{Lester et~al.(2021)Lester, Al-Rfou, and Constant}]{lester2021power}
Brian Lester, Rami Al-Rfou, and Noah Constant. 2021.
\newblock The power of scale for parameter-efficient prompt tuning.
\newblock \emph{arXiv preprint arXiv:2104.08691}.

\bibitem[{Lhoest et~al.(2021)Lhoest, del Moral, von Platen, Wolf, Jernite, Thakur, Tunstall, Patil, Drame, Chaumond, Plu, Davison, Brandeis, Sanh, Scao, Xu, Patry, Liu, McMillan-Major, Schmid, Gugger, Raw, Lesage, Lozhkov, Carrigan, Matussière, von Werra, Debut, Bekman, and Delangue}]{quentin_lhoest_2021_5570305}
Quentin Lhoest, Albert~Villanova del Moral, Patrick von Platen, Thomas Wolf, Yacine Jernite, Abhishek Thakur, Lewis Tunstall, Suraj Patil, Mariama Drame, Julien Chaumond, Julien Plu, Joe Davison, Simon Brandeis, Victor Sanh, Teven~Le Scao, Kevin~Canwen Xu, Nicolas Patry, Steven Liu, Angelina McMillan-Major, Philipp Schmid, Sylvain Gugger, Nathan Raw, Sylvain Lesage, Anton Lozhkov, Matthew Carrigan, Théo Matussière, Leandro von Werra, Lysandre Debut, Stas Bekman, and Clément Delangue. 2021.
\newblock \href {https://doi.org/10.5281/zenodo.5570305} {huggingface/datasets: 1.13.2}.

\bibitem[{Li et~al.(2022)Li, Thickstun, Gulrajani, Liang, and Hashimoto}]{NEURIPS2022_1be5bc25}
Xiang Li, John Thickstun, Ishaan Gulrajani, Percy~S Liang, and Tatsunori~B Hashimoto. 2022.
\newblock \href {https://proceedings.neurips.cc/paper_files/paper/2022/file/1be5bc25d50895ee656b8c2d9eb89d6a-Paper-Conference.pdf} {Diffusion-lm improves controllable text generation}.
\newblock In \emph{Advances in Neural Information Processing Systems}, volume~35, pages 4328--4343. Curran Associates, Inc.

\bibitem[{Li and Liang(2021)}]{li2021prefix}
Xiang~Lisa Li and Percy Liang. 2021.
\newblock Prefix-tuning: Optimizing continuous prompts for generation.
\newblock \emph{arXiv preprint arXiv:2101.00190}.

\bibitem[{Lin(2004)}]{lin-2004-rouge}
Chin-Yew Lin. 2004.
\newblock \href {https://www.aclweb.org/anthology/W04-1013} {{ROUGE}: A package for automatic evaluation of summaries}.
\newblock In \emph{Text Summarization Branches Out}, pages 74--81, Barcelona, Spain. Association for Computational Linguistics.

\bibitem[{Liu and Singh(2004)}]{liu2004conceptnet}
Hugo Liu and Push Singh. 2004.
\newblock Conceptnet—a practical commonsense reasoning tool-kit.
\newblock \emph{BT technology journal}, 22(4):211--226.

\bibitem[{Liu et~al.(2021)Liu, Yuan, Fu, Jiang, Hayashi, and Neubig}]{liu2021pre}
Pengfei Liu, Weizhe Yuan, Jinlan Fu, Zhengbao Jiang, Hiroaki Hayashi, and Graham Neubig. 2021.
\newblock Pre-train, prompt, and predict: A systematic survey of prompting methods in natural language processing.
\newblock \emph{arXiv preprint arXiv:2107.13586}.

\bibitem[{Liu et~al.(2023)Liu, Zheng, Du, Ding, Qian, Yang, and Tang}]{liu2023gpt}
Xiao Liu, Yanan Zheng, Zhengxiao Du, Ming Ding, Yujie Qian, Zhilin Yang, and Jie Tang. 2023.
\newblock \href {http://arxiv.org/abs/2103.10385} {Gpt understands, too}.

\bibitem[{Mostafazadeh et~al.(2016)Mostafazadeh, Chambers, He, Parikh, Batra, Vanderwende, Kohli, and Allen}]{mostafazadeh2016corpus}
Nasrin Mostafazadeh, Nathanael Chambers, Xiaodong He, Devi Parikh, Dhruv Batra, Lucy Vanderwende, Pushmeet Kohli, and James Allen. 2016.
\newblock A corpus and cloze evaluation for deeper understanding of commonsense stories.
\newblock In \emph{Proceedings of the 2016 Conference of the North American Chapter of the Association for Computational Linguistics: Human Language Technologies}, pages 839--849.

\bibitem[{Mostafazadeh et~al.(2020)Mostafazadeh, Kalyanpur, Moon, Buchanan, Berkowitz, Biran, and Chu-Carroll}]{mostafazadeh-etal-2020-glucose}
Nasrin Mostafazadeh, Aditya Kalyanpur, Lori Moon, David Buchanan, Lauren Berkowitz, Or~Biran, and Jennifer Chu-Carroll. 2020.
\newblock \href {https://doi.org/10.18653/v1/2020.emnlp-main.370} {{GLUCOSE}: {G}enera{L}ized and {CO}ntextualized story explanations}.
\newblock In \emph{Proceedings of the 2020 Conference on Empirical Methods in Natural Language Processing (EMNLP)}, pages 4569--4586, Online. Association for Computational Linguistics.

\bibitem[{Peng et~al.(2018)Peng, Ghazvininejad, May, and Knight}]{peng2018towards}
Nanyun Peng, Marjan Ghazvininejad, Jonathan May, and Kevin Knight. 2018.
\newblock Towards controllable story generation.
\newblock In \emph{Proceedings of the First Workshop on Storytelling}, pages 43--49.

\bibitem[{Petroni et~al.(2019)Petroni, Rockt{\"a}schel, Riedel, Lewis, Bakhtin, Wu, and Miller}]{petroni2019language}
Fabio Petroni, Tim Rockt{\"a}schel, Sebastian Riedel, Patrick Lewis, Anton Bakhtin, Yuxiang Wu, and Alexander Miller. 2019.
\newblock \href {https://doi.org/10.18653/v1/D19-1250} {Language models as knowledge bases?}
\newblock In \emph{Proceedings of the 2019 Conference on Empirical Methods in Natural Language Processing and the 9th International Joint Conference on Natural Language Processing (EMNLP-IJCNLP)}, pages 2463--2473, Hong Kong, China. Association for Computational Linguistics.

\bibitem[{Post(2018)}]{post-2018-call}
Matt Post. 2018.
\newblock \href {https://doi.org/10.18653/v1/W18-6319} {A call for clarity in reporting {BLEU} scores}.
\newblock In \emph{Proceedings of the Third Conference on Machine Translation: Research Papers}, pages 186--191, Brussels, Belgium. Association for Computational Linguistics.

\bibitem[{Radford et~al.(2019)Radford, Wu, Child, Luan, Amodei, and Sutskever}]{radford2019language}
Alec Radford, Jeff Wu, Rewon Child, David Luan, Dario Amodei, and Ilya Sutskever. 2019.
\newblock Language models are unsupervised multitask learners.

\bibitem[{Raffel et~al.(2020)Raffel, Shazeer, Roberts, Lee, Narang, Matena, Zhou, Li, and Liu}]{2020t5}
Colin Raffel, Noam Shazeer, Adam Roberts, Katherine Lee, Sharan Narang, Michael Matena, Yanqi Zhou, Wei Li, and Peter~J. Liu. 2020.
\newblock \href {http://jmlr.org/papers/v21/20-074.html} {Exploring the limits of transfer learning with a unified text-to-text transformer}.
\newblock \emph{Journal of Machine Learning Research}, 21(140):1--67.

\bibitem[{Sap et~al.(2019)Sap, Le~Bras, Allaway, Bhagavatula, Lourie, Rashkin, Roof, Smith, and Choi}]{sap2019atomic}
Maarten Sap, Ronan Le~Bras, Emily Allaway, Chandra Bhagavatula, Nicholas Lourie, Hannah Rashkin, Brendan Roof, Noah~A Smith, and Yejin Choi. 2019.
\newblock Atomic: An atlas of machine commonsense for if-then reasoning.
\newblock In \emph{Proceedings of the AAAI Conference on Artificial Intelligence}, volume~33, pages 3027--3035.

\bibitem[{See et~al.(2019)See, Roller, Kiela, and Weston}]{see-etal-2019-makes}
Abigail See, Stephen Roller, Douwe Kiela, and Jason Weston. 2019.
\newblock \href {https://doi.org/10.18653/v1/N19-1170} {What makes a good conversation? how controllable attributes affect human judgments}.
\newblock In \emph{Proceedings of the 2019 Conference of the North {A}merican Chapter of the Association for Computational Linguistics: Human Language Technologies, Volume 1 (Long and Short Papers)}, pages 1702--1723, Minneapolis, Minnesota. Association for Computational Linguistics.

\bibitem[{Shin et~al.(2020)Shin, Razeghi, Logan~IV, Wallace, and Singh}]{shin2020autoprompt}
Taylor Shin, Yasaman Razeghi, Robert~L Logan~IV, Eric Wallace, and Sameer Singh. 2020.
\newblock Autoprompt: Eliciting knowledge from language models with automatically generated prompts.
\newblock \emph{arXiv preprint arXiv:2010.15980}.

\bibitem[{Speer et~al.(2017)Speer, Chin, and Havasi}]{speer2017conceptnet}
Robyn Speer, Joshua Chin, and Catherine Havasi. 2017.
\newblock Conceptnet 5.5: An open multilingual graph of general knowledge.
\newblock In \emph{Thirty-first AAAI conference on artificial intelligence}.

\bibitem[{Sutskever et~al.(2014)Sutskever, Vinyals, and Le}]{sutskever2014sequence}
Ilya Sutskever, Oriol Vinyals, and Quoc~V Le. 2014.
\newblock Sequence to sequence learning with neural networks.
\newblock In \emph{Advances in neural information processing systems}, pages 3104--3112.

\bibitem[{Wei et~al.(2022)Wei, Wang, Schuurmans, Bosma, ichter, Xia, Chi, Le, and Zhou}]{NEURIPS2022_9d560961}
Jason Wei, Xuezhi Wang, Dale Schuurmans, Maarten Bosma, brian ichter, Fei Xia, Ed~Chi, Quoc~V Le, and Denny Zhou. 2022.
\newblock \href {https://proceedings.neurips.cc/paper_files/paper/2022/file/9d5609613524ecf4f15af0f7b31abca4-Paper-Conference.pdf} {Chain-of-thought prompting elicits reasoning in large language models}.
\newblock In \emph{Advances in Neural Information Processing Systems}, volume~35, pages 24824--24837. Curran Associates, Inc.

\bibitem[{Wolf et~al.(2019)Wolf, Debut, Sanh, Chaumond, Delangue, Moi, Cistac, Rault, Louf, Funtowicz et~al.}]{wolf2019huggingface}
Thomas Wolf, Lysandre Debut, Victor Sanh, Julien Chaumond, Clement Delangue, Anthony Moi, Pierric Cistac, Tim Rault, R{\'e}mi Louf, Morgan Funtowicz, et~al. 2019.
\newblock Huggingface's transformers: State-of-the-art natural language processing.
\newblock \emph{arXiv preprint arXiv:1910.03771}.

\bibitem[{Zhang et~al.(2021)Zhang, Geng, Qin, Wu, and Jiang}]{zhangKoala2021}
Zhihan Zhang, Xiubo Geng, Tao Qin, Yunfang Wu, and Daxin Jiang. 2021.
\newblock \href {https://doi.org/10.1145/3442381.3450126} {Knowledge-aware procedural text understanding with multi-stage training}.
\newblock In \emph{Proceedings of the Web Conference 2021}, WWW '21, page 3512–3523, New York, NY, USA. Association for Computing Machinery.

\end{thebibliography}

\appendix

\section{Mechanical Turk Survey}
\begin{figure*}[ht!]
\includegraphics[width=\textwidth]{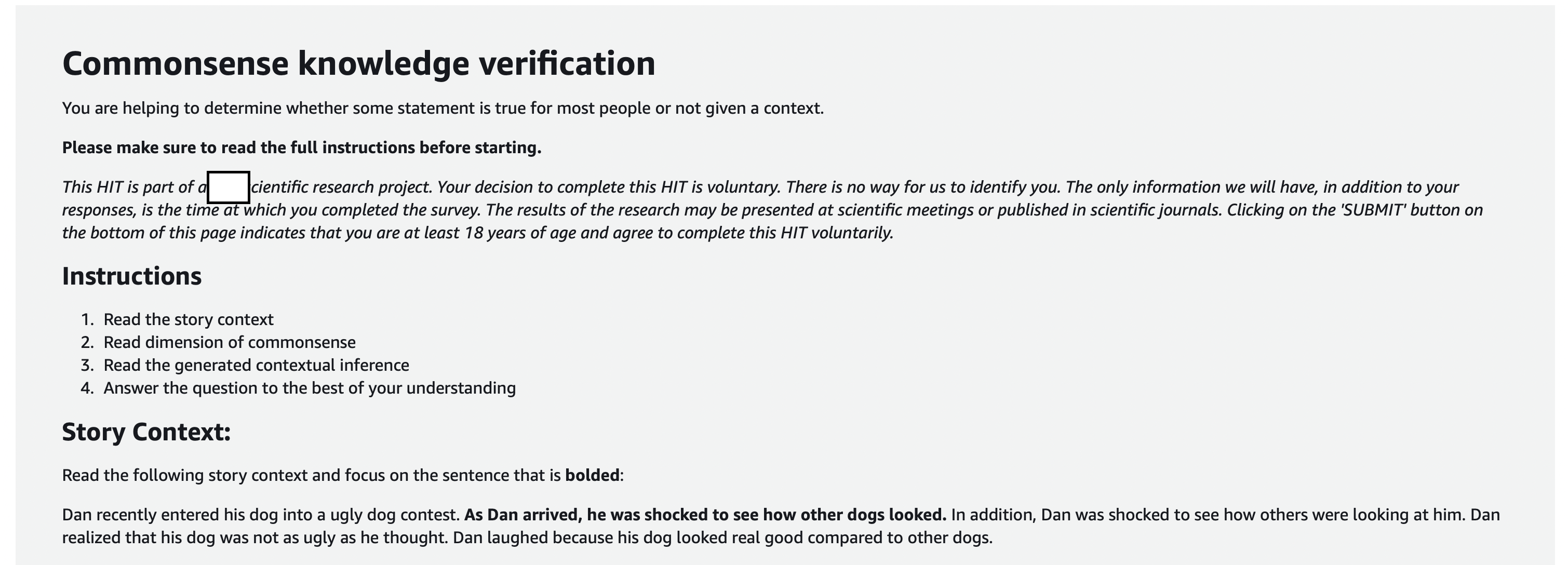}\caption{Screenshot of the Mechanical Turk Task}\label{fig:mechturk1}
\end{figure*}

\begin{figure*}[ht!]
\includegraphics[width=\textwidth]{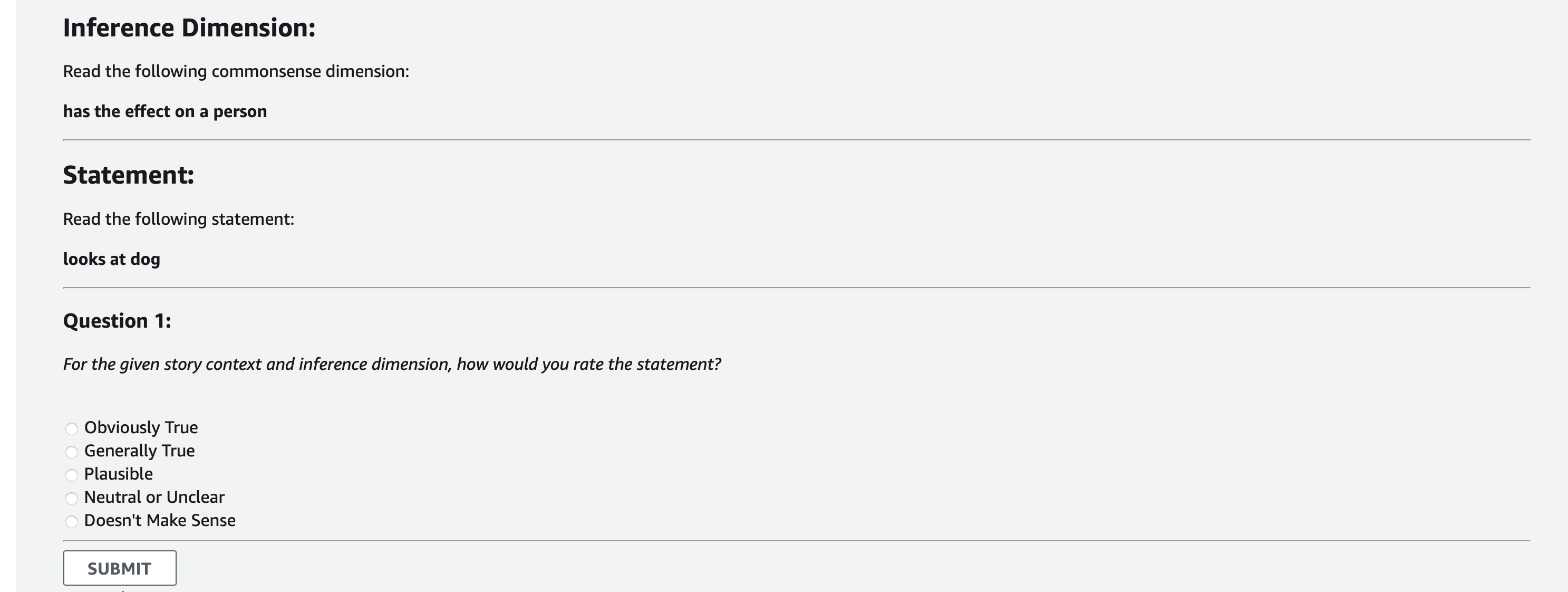}\caption{Screenshot of the Mechanical Turk Task pt.2 }\label{fig:mechturk2}
\end{figure*}

In Figure \ref{fig:mechturk1} and \ref{fig:mechturk2} we can see the survey that was handed to MTurk participants.
\label{sec:appendix1}

\section{Synonym/Antonym Configurations}

The synonym/antonym prompts are added in the same way as the hinting prompts. We experimented with the $p$ value of binomial distribution to control the proportion of synonym/antonym prompts we give during training to look for the most effective proportion. The $p$ value we have used is 0.5. 
An example would be:  \begin{quote}
    \tiny
    \textbf{Synonym + Hinting:} \textit{ The hockey game was tied up. The red team had the puck. They sprinted down the ice. They cracked a shot on goal! They scored a final goal! \textbf{Target sentence:} They scored a final goal! \\\textbf{Hint + synonym prompt:} (\textless{}$|$subj$|$\textgreater{} the red team, \textless{}$|$rel$|$\textgreater{} capable of) (\textless{}$|$subj$|$\textgreater{} the red squad, \textless{}$|$rel$|$\textgreater{} capable of,\textless{}$|$syn$|$\textgreater{})}. 
\end{quote}

\section{ParaCOMET Commonsense Inference Example}
\label{appendix:paracomet_formulation}
An example of an input and expected output from the ParaCOMET formulation can be seen below:
\begin{quote}
\tiny
\textbf{Model Input:} The hockey game was tied up. The {red team} had the puck. They sprinted down the ice. They cracked a shot on goal! \textit{They scored a final goal!} \textless{}$|$sent5$|$\textgreater{} \textless{}$|$xEffect$|$\textgreater{}\\
\textbf{Model Target/Output:}  {win the game}\\
\end{quote}
In this example, since the model is predicting ATOMIC objects, the output is a single phrase (i.e., {win the game}). Additionally, the symbols \textit{\textless{}$|$sent5$|$\textgreater{}} and \textit{\textless{}$|$xEffect$|$\textgreater{} } mean that the target sentence is sentence number five\footnote{We note that in the original ParaCOMET work, the sentences were 0-start indexed. We utilize 1-start indexing for clearer understanding.}, and that the relation we want to generate a tuple about is the "has the effect on a certain person(s)" respectively.  
\section{GLUCOSE Commonsense Inference Example}
\label{appendix:glucose_formulation}
An example of the GLUCOSE formulation's inputs and expected outputs is given below:

\begin{quote}
\tiny
\textbf{Model Input:} 1: The hockey game was tied up. The {red team} had the puck. They sprinted down the ice. They cracked a shot on goal! *\textit{They scored a final goal!}*\\
\textbf{Model Target/Output:} {The red team scores}, {Causes/Enables}, {they win the game} ** {People\_A score}, {Causes/Enables}, {People\_A win a game}\\
\end{quote}

This formulation of contextual commonsense inference is harder than the ParaCOMET one in that it has to generate two sets of {a subject}, {relation}, and {an object} tuples, in which one is the story specific one and the other is the general version of the assertion. These are seen above separated by the ** respectively. In this example additionally, we can see the symbol \textit{1:} which tells the model to predict along a dimension of commonsense described by GLUCOSE (i.e., 1: Event that directly causes or enables X), and the sentence enclosed by asterisks (*) which signifies it is the target sentence.

 \subsection{An example of Hinting with Synonyms and Antonyms}
An example of the synonym/antonym enhanced hinting is the following. Given the following data:
\begin{quote}
    \tiny
    \textbf{Story:} \textit{The hockey game was tied up. The red team had the puck. They sprinted down the ice. They cracked a shot on goal! They scored a final goal!} 
    
    \textbf{Target sentence:} \textit{They scored a final goal! }
    
    \textbf{Target assertion:}
    \textit{(subject: the red team, relation: are capable of, object: winning the game.) }. 
\end{quote}
 A possible hint given to the model and its corresponding synonym and antonym equivalent is: 
 \begin{quote}
    \tiny
    \textbf{Hint:} \textit{(\textless{}$|$subj$|$\textgreater{} the red team, \textless{}$|$rel$|$\textgreater{} capable of) }
    
    \textbf{Synonym Hint:} \textit{(\textless{}$|$subj$|$\textgreater{} the red \textbf{squad}, \textless{}$|$rel$|$\textgreater{} capable of, \textbf{\textless{}$|$syn$|$\textgreater{}}) }. 

    \textbf{Antonym Hint:} \textit{(\textless{}$|$subj$|$\textgreater{} the red \textbf{individual}, \textless{}$|$rel$|$\textgreater{} capable of, \textbf{\textless{}$|$ant$|$\textgreater{}}) }. 

\end{quote}
 Put together with the rest of the story, the input for the model (with a synonym hint) would be: 
 \begin{quote}
    \tiny
 \textbf{Story with Synonym Hint:}  \textit{The hockey game was tied up. The red team had the puck. They sprinted down the ice. They cracked a shot on goal! They scored a final goal! \textbf{(\textless{}$|$subj$|$\textgreater{} the red \underline{squad}, \textless{}$|$rel$|$\textgreater{} capable of, \textbf{\textless{}$|$syn$|$\textgreater{}})}}.
 
\end{quote}
We note that although we provide the example for a synonym, antonyms follow the same process. Additionally, since we sample whether to supply a synonym or antonym, or both, it is possible to give one prompt with all three: hint, synonym hint, and antonym hint.
\section{Experimental Hyperparameter Configurations}
We ran the ParaCOMET experiments for 5 epochs on the dataset's training data and evaluate on the dataset's evaluation data. We utilize a max source sequence length for the T5 models of 256, and a max target length of 128. For the GPT-2 models we utilize a max sequence length of 384.  
Additionally, we use the ADAM \cite{kingma2014adam} optimizer with a learning rate of 2e-5. For the T5 models we utilize a learning rate of 1e-4 because early experiments showed that the model would not converge with lesser learning rates. In both models we use a batch size of 4. We utilize the scripts from \cite{Gabriel_PC} for data generation. The results that we present are the average of the 5 runs over 4 seeds for the conditions.

We ran the GLUCOSE experiments for 5 epochs and 4 seeds on the original  GLUCOSE data. We utilize the ADAM optimizer with a learning rate of 3e-4, a batch size of 8, and a max source length of 256 and max target length of 128. In our results, we present the average of the 4 seeds across the 5 epochs. In both experiments we report the scores given by  SacreBLEU \cite{post-2018-call}, ROUGE \cite{lin-2004-rouge}, and METEOR \cite{banarjee2005} using the datasets library \cite{quentin_lhoest_2021_5570305} metrics system. We run our experiments in a machine with an AMD ThreadRipper 3970 Pro and with NVIDIA A6000s. Every epoch per model is approximately an hour.




\section{Future Work}
\label{appendix:futurework}
When designing the \textit{hinting} system certain aspects were formulated to leave space for improvements. One such area is finding a smarter way of selecting when to \textit{hint}, and finding a smarter way of selecting what to \textit{hint}. Additionally, more soft prompts could be added to the \textit{hint} such that they would learn a better virtual template. 
 
Another area to explore is providing deeper ablation studies to determine what parts of the \textit{hint} are more effective and when.  This work is more a proof-of-concept that \textit{hinting}, or more broadly prompting, is useful towards the task of contextual commonsense inference.  

Exploring further, another approach, chain-of-thought prompting \cite{NEURIPS2022_9d560961}, is commonly utilized in very large language models—a practice that does not extend to the smaller models within our scope. Considering this, we could explore an analogous approach, such as chain-of-hinting, an adaptation that may enable us to incorporate multi-hop graph reasoning for contextual commonsense inference. 

Furthermore, given that models trained with \textit{hinting} for contextual commonsense inference can be guided by the information supplied in \textit{hints}, such models can be utilized in a variety of downstream applications such as story understanding and contextual knowledge graph generation.



\end{document}